%% file: paper.tex
\documentclass[10pt,twocolumn,letterpaper]{article}

\usepackage{iccv}
\usepackage{times}
\usepackage{epsfig}
\usepackage{graphicx}
\usepackage{amsmath}
\usepackage{amssymb}

\usepackage{booktabs}
\usepackage{pifont} %

\usepackage{pgfplots} %
\usepackage{pgfplotstable} %
\pgfplotsset{compat=newest} %
\usetikzlibrary{plotmarks} %

\usepackage[pagebackref=true,breaklinks=true,letterpaper=true,colorlinks,bookmarks=false]{hyperref}

\iccvfinalcopy %

\def\iccvPaperID{1770} %

\newcommand{\PAR}[1]{\vskip1pt \noindent {\bf #1~}}
\newcommand{\PARbegin}[1]{\noindent {\bf #1~}}

\ificcvfinal\fi
\begin{document}

\title{BoLTVOS: Box-Level Tracking for Video Object Segmentation}

\newcommand{\footremember}[2]{%
   \thanks{#2}
    \newcounter{#1}
    \setcounter{#1}{\value{footnote}}%
}
\newcommand{\footrecall}[1]{%
    \footnotemark[\value{#1}]%
} 
\author{Paul Voigtlaender\footremember{cont}{Equal Contribution} \hspace{20pt} Jonathon Luiten\footrecall{cont} \hspace{20pt} Bastian Leibe\\
Computer Vision Group\\
RWTH Aachen University\\
{\tt\small \{voigtlaender,luiten,leibe\}@vision.rwth-aachen.de}
}

\maketitle
\thispagestyle{empty}

\begin{abstract}

We approach video object segmentation (VOS) by splitting the task into two sub-tasks: bounding box level tracking, followed by bounding box segmentation. Following this paradigm, we present BoLTVOS (Box-Level Tracking for VOS), which consists of an R-CNN detector conditioned on the first-frame bounding box to detect the object of interest, a temporal consistency rescoring algorithm, and a Box2Seg network that converts bounding boxes to segmentation masks.
BoLTVOS performs VOS using only the first-frame bounding box without the mask. We evaluate our approach on DAVIS 2017 and YouTube-VOS, and show that it  outperforms all methods that do not perform first-frame fine-tuning.
We further present BoLTVOS-ft, which learns to segment the object in question using the first-frame mask while it is being tracked, without increasing the runtime. BoLTVOS-ft outperforms PReMVOS, the previously best performing VOS method on DAVIS 2016 and YouTube-VOS, while running up to 45 times faster.
Our bounding box tracker also outperforms all previous short-term and long-term trackers on the bounding box level tracking datasets OTB 2015 and LTB35. %
A newer version \cite{Voigtlaender19Arxiv} of this work can be found at \url{https://arxiv.org/abs/1911.12836}.
\end{abstract}

\begin{figure}[t!]
\centering
\includegraphics[width=\linewidth]{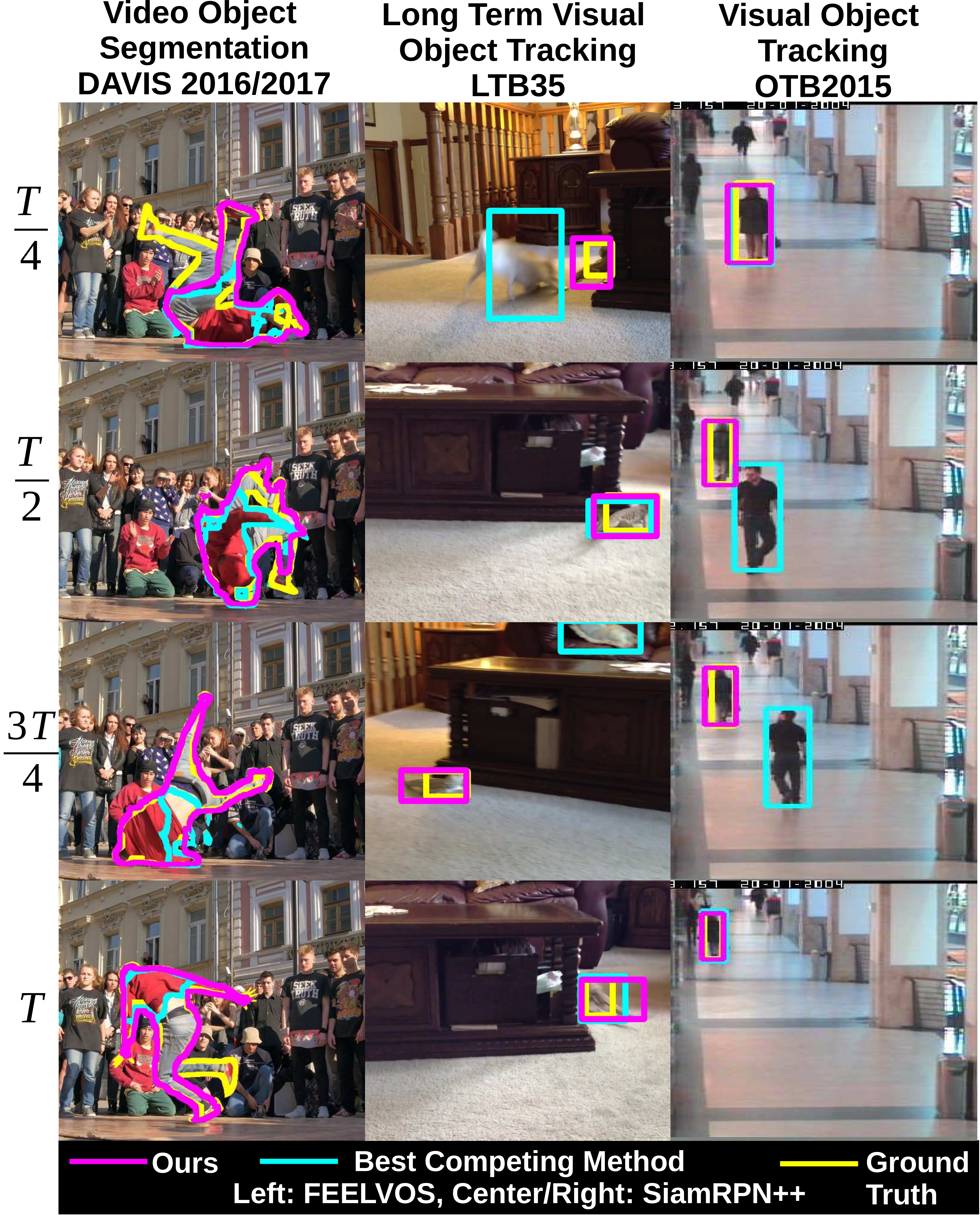}
\vspace{-12pt}
\caption{Example results of BoLTVOS on 3 datasets where it obtains the best results for video object segmentation, visual object tracking and long-term visual object tracking. Our method: purple, ground truth: yellow, previous best method (FEELVOS \cite{Voigtlaender19CVPR} (column 1), SiamRPN++ \cite{Li19CVPR} (columns 2, 3)): blue.}
\label{fig:teaser}
\end{figure}

\vspace{-9pt}
\section{Introduction}
Semi-supervised video object segmentation (VOS) is the task of producing segmentation masks for objects for each frame in a video given the ground truth %
of the first frame. It has important applications including robotics, autonomous driving, and video editing. %
In this paper, we present a novel method for VOS, BoLTVOS (Box-Level Tracking for Video Object Segmentation), which tackles the VOS task using only the first-frame bounding boxes.
 
Current semi-supervised video object segmentation methods belong to one of the following three categories: 1) methods that fine-tune on the first-frame mask, 2) methods that use the first-frame mask without fine-tuning, and 3) methods that only use the first-frame bounding box.

Category 1 methods are extremely slow, but produce very accurate results. Methods in the second category are typically faster but do not perform as well as category 1 methods. Until now, methods in category 3 have performed much worse than methods in the other two categories. BoLTVOS changes this picture and outperforms all category 2 methods without using the first-frame mask.

BoLTVOS is very useful for practical applications such as segmenting objects in large video collections, as it performs well, is fast, and only requires a first-frame bounding box to be annotated rather than the time-consuming process of annotating a first-frame mask. See Fig.~\ref{fig:teaser} for an example of qualitative results of BoLTVOS.

BoLTVOS explores the idea that the VOS task can be divided into 2 sub-tasks. Firstly, the task of tracking the object of interest at a bounding box level, followed by the task of segmenting an object given its bounding box.
In order to tackle the first sub-task, we introduce a novel conditional R-CNN network followed by a temporal consistency rescoring step to improve the tracking results. We address the second sub-task by generating segmentations for each bounding box using the Box2Seg network from \cite{Luiten18ACCV}.

Many recent visual object tracking approaches \cite{Li18CVPRSiamRPN,Zhu18ECCV,Li19CVPR} use a single stage object detector that is conditioned by performing a cross-correlation between template features and the features of a local search region around the previous prediction \cite{Li18CVPRSiamRPN}.
In contrast, our conditional R-CNN evaluates possible regions over the whole image by using a two stage R-CNN style network with a novel conditional second stage for tracking, which directly compares template features to region proposal features by concatenating the features before predicting their similarity.
This allows our method to recover from incorrect detection results, and to re-detect an object after disappearance.
We further improve our tracking method by adding a temporal consistency rescoring algorithm, which rescores detection results by taking into account temporal consistency cues while also modeling potential distractor objects, using tracklet-based online dynamic programming.

We evaluate BoLTVOS on three VOS benchmarks, DAVIS 2016 \cite{DAVIS2016}, DAVIS 2017 \cite{DAVIS2017}, and YouTube-VOS \cite{Xu18ECCV}. On all three datasets, BoLTVOS performs much better than all other methods that also only use the first-frame bounding box. %
Our method also outperforms all other category 2 VOS methods, and many category 1 VOS methods on DAVIS 2017 \cite{DAVIS2017} and YouTube-VOS \cite{Xu18ECCV}, even though it does not use the first-frame mask.

For scenarios where the first-frame ground truth mask is available, we propose an extension, BoLTVOS-ft, which fine-tunes Box2Seg on the first-frame mask in parallel to tracking. Although this extension uses fine-tuning, it is as fast as methods that do not. On DAVIS 2016 \cite{DAVIS2016} and YouTube-VOS \cite{Xu18ECCV}, BoLTVOS-ft outperforms all other methods, while running up to 45 times faster than the previous best method.
We additionally evaluate the bounding box tracking performance of BoLTVOS separately on two visual object tracking benchmarks, where we also achieve new state-of-the-art results.

\section{Related Work}
\PARbegin{Video Object Segmentation (VOS).}
VOS methods can be divided into three categories. Category 1 methods perform fine-tuning on the first-frame ground truth masks \cite{OSVOS,Maninis18TPAMI,voigtlaender17BMVC,Li18ECCV,Bao18CVPR,Luiten18ACCV}, which leads to impressive results but is slow.

Category 2 methods use the first-frame mask without fine-tuning \cite{Chen18CVPR,Yang18CVPR,Cheng18CVPR,Hu18ECCV,Oh18CVPR,Xu18ECCV,Voigtlaender19CVPR}. These methods run much faster, but do not achieve the same result quality.

The third category of VOS methods are those that do not use the first-frame mask at all but are only conditioned on the first-frame bounding box. Our method falls into this category. The only other published method that adopts this approach is SiamMask \cite{Wang19CVPR}. In this regard SiamMask \cite{Wang19CVPR} is the work that is most closely related to ours. Like our method, SiamMask is a box-level tracking method that also produces segmentation masks. Unlike our method, SiamMask only uses a single stage for detection and is only evaluated in a local search window of a previous prediction.

Like BoLTVOS and SiamMask \cite{Wang19CVPR}, also RGMP \cite{Oh18CVPR} adopts a Siamese architecture for VOS. However, RGMP requires the first-frame mask, while BoLTVOS can work with only a bounding box and still achieves better results. Additionally, RGMP performs segmentation on the whole image, whereas we only segment bounding box regions.

We treat the generation of segmentation masks as a post-processing step which happens after the bounding box level tracking. Producing segmentation masks conditioned on a bounding box for VOS has been explored by PReMVOS \cite{Luiten18ACCV} whose segmentation network we adopt. Except for that, PReMVOS is quite different from our method. PReMVOS uses four different neural networks, is very slow, and uses significant fine-tuning.

\begin{figure*}[t]
\centering
\includegraphics[width=\linewidth]{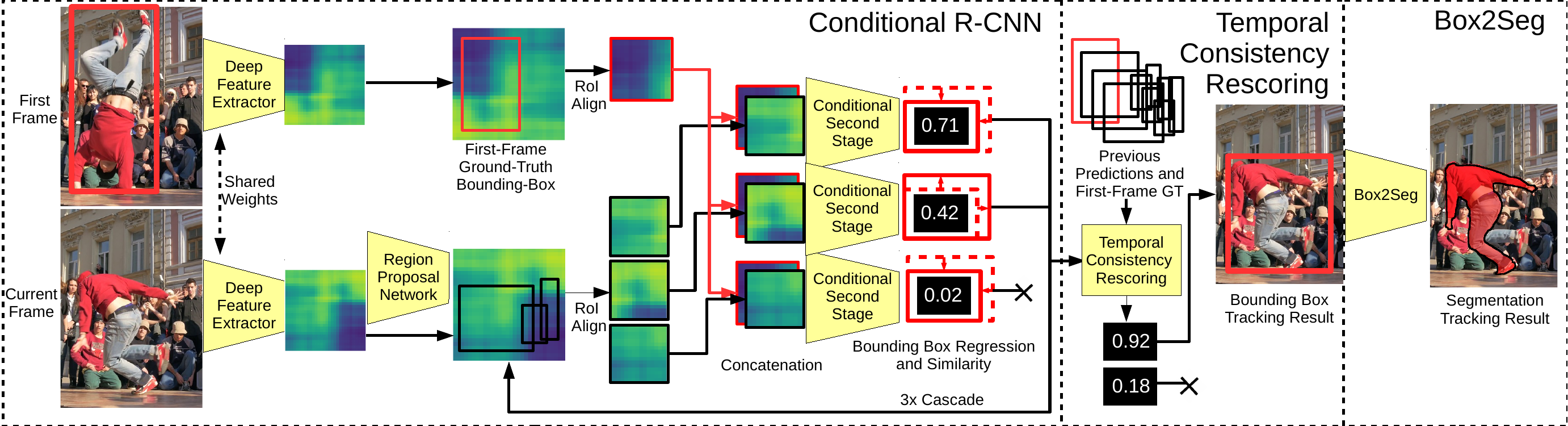}
\caption{Overview of BoLTVOS. A conditional R-CNN (left) provides detections conditioned on the first-frame bounding box, which are then rescored by our temporal consistency rescoring algorithm (center). The result are bounding box level tracks which are converted to segmentation masks by the Box2Seg network (right).}
\label{fig:overview}
\end{figure*}

\PARbegin{Visual Object Tracking (VOT).}
BoLTVOS splits the VOS task into box-level tracking and bounding box segmentation. Box-level tracking is commonly evaluated as the VOT task, which is similar to VOS, but only requires bounding box input and output instead of segmentation masks, and only deals with a single object.
Recently, a number of long-term VOT benchmarks have been released which differ from traditional VOT in that the object to be tracked often  disappears and reappears. %
The field of VOT has been driven by the success of a number of recent benchmarks such as the yearly VOT challenges \cite{Kristan16TPAMI,Kristan18ECCVW}, the Online Tracking Benchmark (OTB) \cite{Wu13CVPR,Wu15TPAMI}, LTB35 \cite{Kristan18ECCVW} and OxUvA \cite{Valmadre18ECCV} for long-term VOT, and many others \cite{Muller18ECCV,Huang18Arxiv,Zhu18Arxiv,Mueller16ECCV,Fan19CVPRLASOT}.

The proposed BoLTVOS box-level tracker works by adapting an object detector for tracking by conditioning the detection on the given first-frame object template. Many previous methods also approach tracking as conditional detection. Most notably Siamese region proposal network methods such as SiamRPN \cite{Li18CVPRSiamRPN} which work by using a single stage RPN \cite{Ren15NIPS} object detector and conditioning it on the first frame by cross-correlating the deep features of a local image patch with the deep features of the template.

There have been many recent improvements of the SiamRPN approach such as DaSiamRPN \cite{Zhu18ECCV} (distractor aware), C-RPN \cite{Fan19CVPRCRPN} (cascaded RPN), SiamMask \cite{Wang19CVPR}, SiamRPN+ \cite{Zhang19CVPR}, and SiamRPN++ \cite{Li19CVPR} (deeper network architectures). All of these methods work by searching within a local window of the previous prediction using a template cross-correlation \cite{Bolme10CVPR,HenriquesTPAMI15,MaICCV15} followed by an RPN detector.

Long-term tracking with Siamese trackers is so far mainly addressed by enlarging the search window when the detection confidence is low \cite{Zhu18ECCV, Li19CVPR}.
Our approach is able to take advantage of global search over the whole image to produce even stronger results than the previous best method SiamRPN++ \cite{Li19CVPR} over all benchmarks, but especially on the long-term VOT benchmark where our method excels. 

\section{Method}
BoLTVOS explores the key idea that the task of video object segmentation (VOS) can be tackled by splitting it into first tracking on a bounding box level, followed by segmenting the objects given by the tracked bounding boxes. By adopting this paradigm, we can draw inspiration from the visual object tracking (VOT) community for designing a box-level tracker that works well for VOS. 

As shown in Fig.~\ref{fig:overview}, BoLTVOS consists of three components. The left figure region shows our conditional Siamese cascaded R-CNN detector that is able to detect object regions that are visually similar to the given first-frame template object. The center figure region shows our online temporal consistency rescoring algorithm that is able to choose the best detection that comes from our detector in each time step based on temporal consistency and visual similarity cues, taking into account the potential presence of visually similar, but spatially inconsistent distractor objects. Finally, as shown in the right figure region, after determining the bounding box level tracking result, we apply our Box2Seg network to generate a segmentation mask for the object given by the bounding box each time step.

Inspired by the success of Siamese networks in VOT benchmarks \cite{Kristan18ECCVW,Wu15TPAMI,Kristan16TPAMI}, we adopt a Siamese style architecture for our conditional detector. The most commonly used VOT benchmarks \cite{Wu15TPAMI,Kristan16TPAMI} only require short-term tracking without object disappearance and reappearance. %
With this domain in mind, most current Siamese tracking architectures \cite{Zhu18ECCV,Wang19CVPR,Li19CVPR,Li18CVPRSiamRPN} employ a strong temporal consistency constraint on their detections, by only searching for new detections within a small spatial window of a previous detection. However, VOS, like the recently introduced long-term VOT task \cite{Kristan18ECCVW,Valmadre18ECCV}, requires the tracking of objects that can disappear and reappear again. In order to tackle this long-term tracking task, we develop a tracking architecture that first detects a number of bounding box proposals using only visual similarity and that then rescores them in a second step using temporal consistency cues. This way, BoLTVOS effectively leverages both visual and temporal cues and is able to re-detect objects after they disappear.%

The previous best-performing bounding box tracking methods \cite{Zhu18ECCV,Wang19CVPR,Li19CVPR,Li18CVPRSiamRPN} adopt a single shot detector, which is conditioned on the first-frame template using cross-correlation. For the task of single-image object detection, two-stage detector networks such as Faster R-CNN \cite{Ren15NIPS} have been shown to outperform single-stage detectors. Inspired by this, we design our tracker as a conditional two-stage detection network. In addition to the increased performance of two-stage detection over single-stage detection, this also brings the advantage that, using Region of Interest (RoI) proposals as an input to the second stage, we can directly compare a proposed RoI region to a template region by concatenating their RoI aligned feature representations instead of performing a cross-correlation. This direct comparison can more effectively learn the similarity of two regions and can also deal with changes in object size and aspect ratio as the proposals are aligned to the same size.

\PAR{Conditional R-CNN.}
The first stage of BoLTVOS is a conditional detector used to detect bounding box regions that are visually similar to the first-frame template (see Fig.~\ref{fig:overview} left). We base our architecture on the two-stage detection architecture of Faster R-CNN \cite{Ren15NIPS}. Specifically, we take a Faster R-CNN architecture that has been pre-trained for single image object detection on the COCO \cite{coco} dataset for detecting 80 object classes. Such a network consists of a backbone feature extractor followed by two detection stages; first a category-agnostic RPN, followed by a category-specific second stage. We fix the weights of the backbone and the RPN and replace the category-specific second stage with our conditional second stage. 

Our second stage is run for each region proposed by the RPN by performing RoI Align \cite{He17ICCV} to extract deep features from this proposed region. We also take the RoI Aligned deep features of the ground truth bounding box in the first-frame image, and then concatenate these together with the features of the proposed region and feed the combined features into a $1\times1$ convolution which reduces the number of features channels back down by half. 
These conditional features are then fed into the Faster R-CNN second stage with two output classes; either the proposed region is the object to be detected or it is not. 
Our conditional detector uses a 3-stage cascade \cite{Cai18CVPR} without shared weights. Using a cascade improves the results, as each cascade layer is trained on the output of the layers below it. This means that with each cascade layer, it is trained with harder and harder negatives and learns to better discriminate between positives and hard negative distractors.
The backbone and RPN are frozen having been trained for detection on COCO. 
Only the second stage (after concatenation) is trained for tracking, using pairs of frames from video datasets. Here, an object in one frame is used as reference and the network is trained to detect the same object in another frame.

\PAR{Temporal Consistency Rescoring.} 
After detecting regions that are visually similar to the first-frame template with our conditional R-CNN, BoLTVOS uses a temporal consistency rescoring algorithm (see Fig.~\ref{fig:overview} center) to rescore the detections. %
This algorithm works in an online manner by first grouping all previous detections from a video into a set of tracklets that are very likely to be the same object \cite{HuangECCV08}. It then scores each tracklet based on its component detection scores, as well as a long-term temporal consistency score measuring the likelihood that this tracklet is a continuation from the object to be tracked in the first frame.

Specifically, tracklets are created in an online manner by adding a detection to an existing tracklet each frame if it has an IoU with the last detection of a tracklet greater than a threshold (around 70\%). Each detection that does not join an existing tracklet creates a new tracklet.
Our algorithm then finds the optimal set of tracklets which make up the final tracking result. It does this by scoring a number of 'track hypotheses' \cite{Osep19arxiv}, different combinations of tracklets, using an online dynamic programming formalization. Tracklets that have overlapping time-steps cannot be composed together.

Formally, let $b_{\mathit{ff}}$ denote the first-frame bounding box of the object to be tracked. A track hypothesis $\mathcal{T}=\{\tau_1,\dots,\tau_k\}$ consists of $k$ tracklets where each tracklet $\tau_i$ has a corresponding start time $t_i^{start}$ and end time $t_i^{end}$ such that tracklet $\tau_i$ starts after tracklet $\tau_{i-1}$ has ended, potentially with a gap between them. Each tracklet consists of detections for each of its time steps with a corresponding bounding box and detection score provided by our conditional R-CNN, \ie, $\tau_i=\{d_{i,t_i^{start}}, \dots, d_{i,t_i^{end}}\}$, where $d_{i,t}=(b_{i,t}, s_{i,t})$ and $b_{i,t}$ is the bounding box of detection $d_{i,t}$ of tracklet $\tau_i$ at time $t$ and $s_{i,t}$ is its detection confidence.
The score of a tracklet $\tau_i$ is given by
\vspace{-8pt}
\begin{equation}
\mathrm{score}(\tau_i) = \sum_{t=t_i^{start}}^{t_i^{end}} s_{i,t} + w_{\textit{ff}} \mathrm{ff\_score}(b_{\textit{ff}}, b_{i,t}),
\end{equation}
\vspace{-11pt}
\begin{equation}
\mathrm{ff\_score}(b_{\textit{ff}}, b) = \min(\frac{ar(b_{\textit{ff}})}{ar(b)}, \frac{ar(b)}{ar(b_{\textit{ff}})}) - \alpha_{\textit{ff}}.
\end{equation}
\vspace{-2pt}
Here, $\mathrm{ff\_score}$ measures the similarity in aspect ratio between the current detection and the first-frame bounding box and $ar(b)=\frac{width(b)}{height(b)}$ denotes the aspect ratio of bounding box $b$, and $w_{\textit{ff}}$ and $\alpha_{\textit{ff}}$ are hyperparameters.

The score of a whole track hypothesis $\mathcal{T}=\{\tau_1,\dots,\tau_k\}$ is a combination of the scores of its tracklets and boundary scores between successive tracklets, \ie,
\vspace{-8pt}
\begin{equation}
\mathrm{score}(\mathcal{T}) = \sum_{i=1}^{k} \mathrm{score}(\tau_i) +  w_{bnd} \sum_{i=2}^k \mathrm{bnd\_score}(\tau_{i-1}, \tau_i)
\end{equation}
\vspace{-10pt}
\begin{equation}
\begin{split}
\mathrm{bnd\_score}(\tau_{i-1}, \tau_i) = \sum_{t=t_{i-1}^{start}}^{t_{i-1}^{end}} \big( w_{iou} \mathrm{IoU}(b_{i-1,t_{i-1}^{end}}, b_{i,t_{i}^{start}})\\
-w_{loc} \lVert \mathrm{center}(b_{i-1,t_{i-1}^{end}}) - \mathrm{center}(b_{i,t_{i}^{start}}) \rVert - \alpha_{bnd} \big).
\end{split}
\end{equation}
Here, $\lVert \cdot \rVert$ denotes Euclidean distance, $\mathrm{center}$ is the center of a bounding box in pixel coordinates, $w_{iou}, w_{loc}$, and $\alpha_{bnd}$ are hyperparameters, and the IoU term calculates the spatial intersection-over-union between the last bounding box of tracklet $\tau_{i-1}$ and the first bounding box of tracklet $\tau_i$, regardless of a potential temporal gap.
Using dynamic programming and only keeping the highest score track hypothesis for each tracklet results in a very small and manageable set of track hypotheses. Both tracklet generation and track hypothesis scoring can be performed iteratively online with only a small set of tracklets and track hypotheses updated with each new time-step. We then choose the highest-scoring track hypothesis and output the detection from the most recent time-step that belongs to this track hypothesis. If no detections in the current frame belong to the highest-scoring track hypothesis, we choose a detection in the current time-step based on its score and its temporal consistency to the last detection in our chosen track hypothesis.

Thus, we model a number of different track hypotheses, even those that are not temporally consistent with the first frame bounding box, and we are able to down-score detections that belong to such tracks even if they are extremely visually similar to the object to be tracked.

\PAR{Box2Seg.}
In order to produce segmentation masks for the VOS task, we use an off-the-shelf bounding-box-to-segmentation-mask network by adopting the code and pre-trained weights from Luiten \etal~\cite{Luiten18ACCV}. This network is a fully convolutional DeepLabV3+ \cite{Chen18ECCV} network with an Xception-65 \cite{CholletCVPR17} backbone. It has been trained on Mapillary \cite{Neuhold17ICCV} and then COCO \cite{coco} to output a segmentation mask given by the object bounding box encoded as a fourth input channel. 
This network runs much faster than our conditional R-CNN and is able to convert 40 bounding boxes to segmentation masks per second. Thus, running this network after our tracking method only increases the runtime by 0.025 seconds per object per frame.
We combine overlapping masks such that the mask with the smallest number of pixels ends up on top of other masks.

\PAR{BoLTVOS-ft.} BoLTVOS performs video object segmentation while only using the first-frame bounding box. We present an extension, BoLTVOS-ft, which is able to use first-frame mask annotations, if present, to improve the results. In this extension, the Box2Seg network is fine-tuned on the first-frame object mask. %
Because the Box2Seg network is only run after box-level tracking is complete, the fine-tuning can run in parallel to the tracking, improving the segmentation accuracy without increasing the runtime.

\PAR{Implementation Details.} For our conditional R-CNN we adapt a Faster R-CNN \cite{Ren15NIPS} implementation with a ResNet101 \cite{resnet} backbone with a feature pyramid network \cite{Lin17CVPR}. This has been trained from scratch \cite{He18Arxiv} (without ImageNet \cite{imagenet} pre-training) on COCO \cite{coco} for 80 class object detection and instance segmentation using the mask head from Mask R-CNN \cite{He17ICCV}, group normalization \cite{Wu18ECCV}, a cascaded second stage \cite{Cai18CVPR}.
We use the implementation and pre-trained weights from \cite{Wu16tensorpack}.

As we do not train our RPN specifically for tracking (it generates generic object proposals), we need to generate a larger number of region proposals than used for conventional two-stage detectors to feed into the second stage, but we can use strong non-maximum suppression to keep the number of proposals manageable. 

We train our conditional R-CNN on multiple tracking datasets simultaneously: %
 ImageNet VID \cite{imagenet} (4000 videos), YouTube-VOS \cite{Xu18ECCV} (3471 videos), GOT-10k \cite{Huang18Arxiv} (9335 videos) and the YouTube-BoundingBoxes \cite{Real17CVPR} validation set (around 30000 videos). We sample each of the other datasets twice as often as YouTube-BoundingBoxes to reduce the dataset size bias. We train with motion blur and grayscale augmentations  as done in \cite{Zhu18ECCV}, as well as gamma and scale augmentations.
We train our network on a computer with four 1080 Ti GPUs for 4 days. BoLTVOS is evaluated on a computer with a single V100 GPU, while BoLTVOS-ft uses two V100 GPUs in parallel.

\section{Experiments}
We split the experimental evaluation into three sections. 
First, we evaluate our method on multiple VOS benchmarks.
Then, we evaluate BoLTVOS for long-term visual object tracking, and lastly we evaluate it for standard (short-term) online visual object tracking.

\subsection{Video Object Segmentation Evaluation}
\PARbegin{DAVIS 2017.} We perform experiments on the standard VOS dataset DAVIS 2017 \cite{DAVIS2017} and present results on the validation set, which contains 30 videos with an average of 2.03 objects per video, and a maximum of 5 objects to be tracked in a single video.
The first-frame ground truth segmentation masks are given and the objects must be tracked and segmented throughout the remaining frames.

We adopt the standard metrics for evaluating on DAVIS 2017, the $\mathcal{J}$ score being the average IoU between the predicted mask and the ground truth mask, and the $\mathcal{F}$ score  measuring the similarity of the boundary of both masks.
Methods are ranked on DAVIS 2017 by the average of their $\mathcal{J}$ and $\mathcal{F}$ scores, which is called the $\mathcal{J} \& \mathcal{F}$ measure. Additionally, we introduce a new metric $\mathcal{J}_{box}$, which is analogous to the $\mathcal{J}$ metric for segmentation masks (average IoU), except that it is calculated using the bounding box surrounding a given predicted and ground truth mask, rather than the actual mask. Using both the mask metrics and our new box metric, we are able to evaluate our bounding box tracking method directly, as well as to separately evaluate the effect of predicting segmentation masks.

We run our method on each object separately without any interaction between the runs of separate objects. This means that if the predicted masks overlap, only one is chosen per pixel and thus, if our tracker predicts overlapping regions, our method could be penalized heavily. Hence, this is a good evaluation of the performance of running a single-object tracker on multiple objects in the same video, as it penalizes the overlap of predictions.

We tune our hyperparameters on the DAVIS 2017 training set. Table \ref{tab:results-davis17} shows the results of our evaluation of both the standard BoLTVOS (Ours), without temporal consistency rescoring (Ours (No Rescoring)), and our fine-tuned extension (Ours (Fine-tun.~Box2Seg)) on the validation set and compares it to 12 other state-of-the-art VOS approaches. These approaches are divided into three groups, the first are those that only use the first-frame bounding box, without using the mask. This includes our method. The second group are those that use the first-frame mask but do not fine-tune on it. The third set of methods perform slow fine-tuning on the first-frame mask.
\input{tables/davis-val.tex}

Our method achieves a $\mathcal{J} \& \mathcal{F}$ score of $71.9\%$. This significantly outperforms SiamMask \cite{Wang19CVPR}, the previous best method that only uses the first-frame bounding boxes, with a $\mathcal{J} \& \mathcal{F}$ score of $55.8\%$ ($+16.1\%$). In order to verify that this improvement  does not just stem from a better way to generate masks, we applied the Box2Seg mask generation on the results of SiamMask as a post-processing step. This does improve the results to $63.3\% \mathcal{J} \& \mathcal{F}$, which is still $8.6$ percentage points lower than our result.
Our method also outperforms all of the current methods that use the first-frame mask (without fine-tuning), even though our method only has access to the first-frame bounding box. The previous best performing method of this category is FEELVOS \cite{Voigtlaender19CVPR} which BoLTVOS outperforms by $0.4\%$.

The no-rescoring variant of our method only reaches $64.9\%$ ($-7.0\%$), highlighting the importance of temporal consistency rescoring. The reason for this is that without rescoring, our tracker often produces overlapping predictions, which are then removed, as for the VOS evaluation each pixel can only be a part of a single object. However, when using our rescoring algorithm, we are able to take into account both temporal consistency and the presence of distractor objects in the video, and as such the results are much less likely to overlap resulting in much improved results.

\input{tables/speedplot.tex}
\input{tables/speedplot_ytbvos}

Figure \ref{fig:qual_vs_time} compares our method with other state-of-the-art methods in VOS performance as well as runtime. It can be seen that BoLTVOS outperforms every previous method except PReMVOS \cite{Luiten18ACCV} and DyeNet \cite{Li18ECCV}, which both use slow first-frame fine-tuning. BoLTVOS even significantly outperforms 4 methods that do use slow first-frame fine-tuning. 
Compared to all other methods that use the first-frame mask without fine-tuning, not only does our method perform better, it is also much more practical in real world scenarios, such as segmenting a large collection of videos. This is because first-frame bounding boxes are easy and cheap to annotate, whereas segmenting accurate first-frame masks is a very difficult and expensive procedure.

On the $\mathcal{J}_{box}$ metric, BoLTVOS gets a score of $78.5\%$. This significantly outperforms all other methods except again for PReMVOS \cite{Luiten18ACCV}, where our method is only $2.9\%$ worse, even though the $\mathcal{J} \& \mathcal{F}$ score is $5.9\%$ worse. When comparing to FEELVOS \cite{Voigtlaender19CVPR}, our $\mathcal{J} \& \mathcal{F}$ score is only just higher and FEELVOS has a $\mathcal{J}$ score that is $0.7\%$ higher than ours. However, when looking at the $\mathcal{J}_{box}$ score our method significantly outperforms FEELVOS by $7.1\%$. Comparing to SiamMask, which also performs box-level tracking, BoLTVOS has a $\mathcal{J}_{box}$ score 14.2 percentage points higher. This indicates that our BoLTVOS box-level tracking is performing extremely strongly and that most of the loss in score comes from the out-of-the-box Box2Seg network. 
The loss in performance by Box2Seg can also be seen when applying it to the perfect ground truth bounding boxes, in which case $\mathcal{J}_{box}$ by definition is $100\%$, but the $\mathcal{J} \& \mathcal{F}$ score is only $82.6\%$.

We also perform an experiment where we fine-tune the Box2Seg network while leaving the bounding box tracking component unchanged (Ours (fine-tun.~Box2Seg)). Because Box2Seg is only applied as a post-processing step, it can be fine-tuned in parallel to running the slower conditional R-CNN, and we can then apply the fine-tuned Box2Seg on the box tracking result. We restrict the number of fine-tuning steps to 300, which significantly improves the results but is still fast enough to finish before the conditional R-CNN processes the entire sequence. 
In this way, if a second GPU is available, BoLTVOS-ft produces a final segmentation result without any additional run-time cost.
In this setup, BoLTVOS-ft achieves a $\mathcal{J} \& \mathcal{F}$ score of $76.3\%$, which is very close to the result of PReMVOS with $77.8\%$. Note that PReMVOS is 25 times slower than BoLTVOS-ft and uses 4 different neural networks.

\PAR{DAVIS 2016.} We also present results on the DAVIS 2016 \cite{DAVIS2016} benchmark. %
This benchmark has 20 videos, each with a single object, and uses the same evaluation metrics.

\input{tables/davis16.tex}

Table~\ref{tab:results-davis16} shows the results of BoLTVOS both with (Ours), and without (Ours (No Rescoring)) temporal consistency rescoring and compares it to the same 12 other state-of-the-art VOS approaches, as well as to PML \cite{Chen18CVPR} that does not have DAVIS 2017 results. These methods are divided into the same three groups as with DAVIS 2017: those that do not use the first-frame mask; those that use it without fine-tuning on it; and those that fine-tune on the first-frame mask.

BoLTVOS achieves a $\mathcal{J} \& \mathcal{F}$ score of $79.6\%$. This again significantly outperforms SiamMask \cite{Wang19CVPR}, the previous best method that only uses the first-frame bounding boxes, with $69.8\%$ $\mathcal{J} \& \mathcal{F}$ ($+9.8\%$). Our method does not perform as well on the $\mathcal{J} \& \mathcal{F}$ metric as some of the other methods that do use the first-frame mask. Specifically, FEELVOS \cite{Voigtlaender19CVPR} and RGMP \cite{Oh18CVPR} outperform BoLTVOS by $2.2\%$ and $2.1\%$, respectively. However, when we look at the $\mathcal{J}_{box}$ metric, BoLTVOS outperforms FEELVOS by $8.1\%$ and RGMP by $9.0\%$. This indicates that our tracking method is able to very successfully track the objects in the DAVIS 2016 sequences, but that the out-of-the-box Box2Seg network fails to give us accurate segmentations for this dataset. This is because Box2Seg was only trained on single objects (like a bike, or a person), whereas the objects to be segmented in DAVIS 2016 are often a grouping of multiple objects (such as a person riding a bike).
When using the fine-tuned version of Box2Seg (Ours (Fine-tun.~Box2Seg)), we achieve a $\mathcal{J} \& \mathcal{F}$ score of $87.7\%$, which is higher than the previous best result achieved by PReMVOS with $86.8\%$ (which is 45 times slower than BoLTVOS-ft here).

In this experiment, the effect of our rescoring algorithm is negligible, giving us no boost in performance. This is because in the DAVIS 2016 sequences there is only one main object visible, and there are almost no distractor objects. Hence, using a distractor aware temporal consistency rescoring algorithm is not able to bring any gains, whereas it helped a lot on the much harder DAVIS 2017 dataset.

\PAR{YouTube-VOS.} Finally, we present results on the much larger and more challenging YouTube-VOS dataset \cite{Xu18ECCV}. Here, we re-use the same hyperparameters which we used for DAVIS 2017 obtained by tuning on the training set. 
For YouTube-VOS, the evaluation distinguishes between object classes which are part of the training set (\textit{seen}) and those that are not (\textit{unseen}). The primary evaluation measure $\mathcal{J} \& \mathcal{F}_{seen+uns.}$ is the average of the $\mathcal{J} \& \mathcal{F}$ scores for seen and unseen object classes. 

Table~\ref{tab:results-youtubevos} and Fig.~\ref{fig:speedplot-ytbvos} compare the results and speed of BoLTVOS to other methods on the validation set consisting of 474 sequences. Note that YouTube-VOS is a very recent dataset and hence only few methods have been evaluated on it. The only method which achieves better results on this dataset is PReMVOS \cite{Luiten18ACCV} with a $\mathcal{J} \& \mathcal{F}_{seen+uns.}$ score of $66.9\%$ which is only $1.2\%$ higher than BoLTVOS's result, although PReMVOS uses the ground truth masks of the first frame and performs slow fine-tuning. When we add fine-tuning for Box2Seg, BoLTVOS-ft achieves a score of $71.1\%$,  which sets a new state-of-the-art for this dataset.

\input{tables/youtubevos.tex}

\begin{figure}[t]
 \scalebox{1.02}{
\includegraphics[scale=0.32]{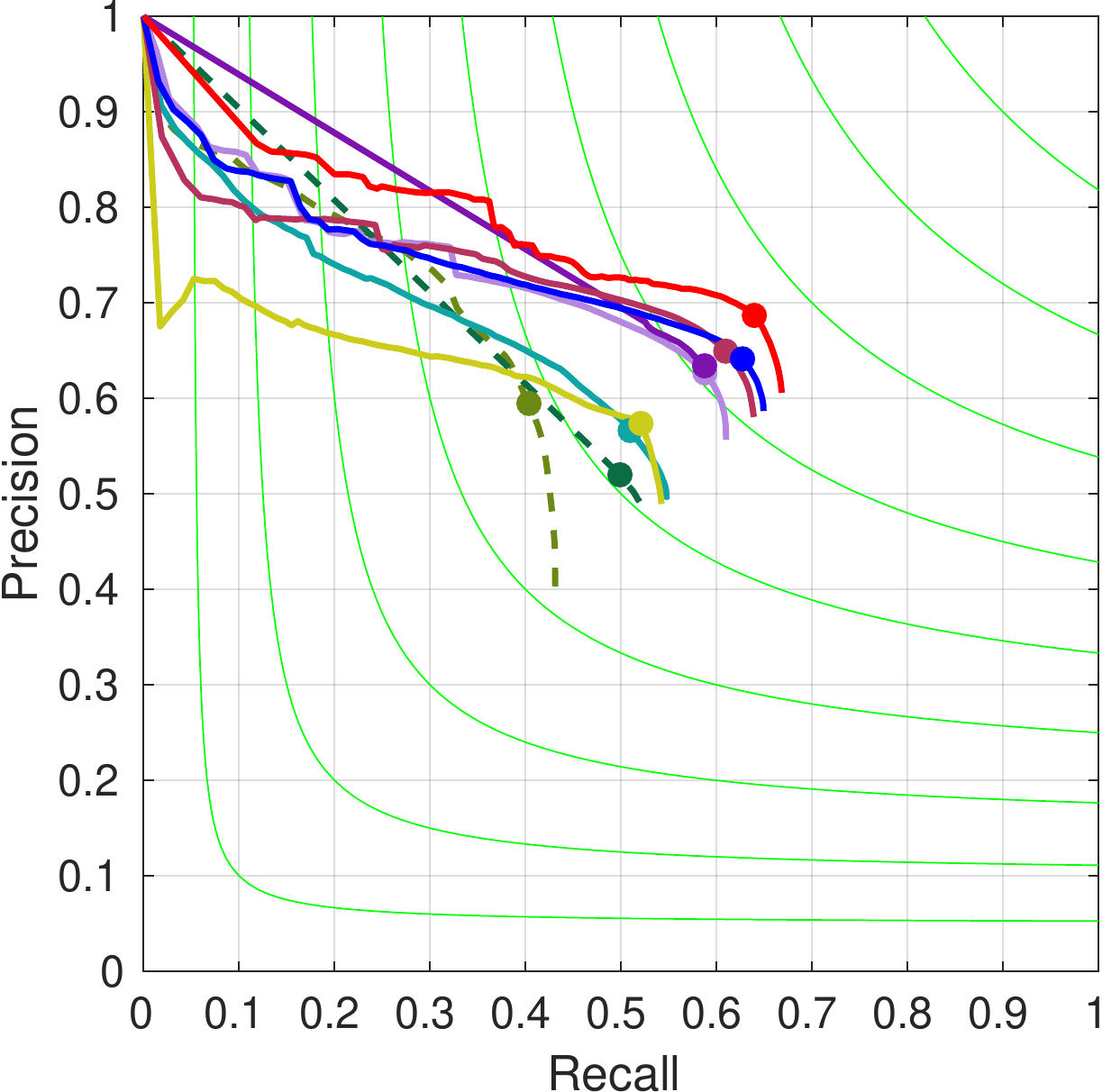}
\includegraphics[scale=0.32]{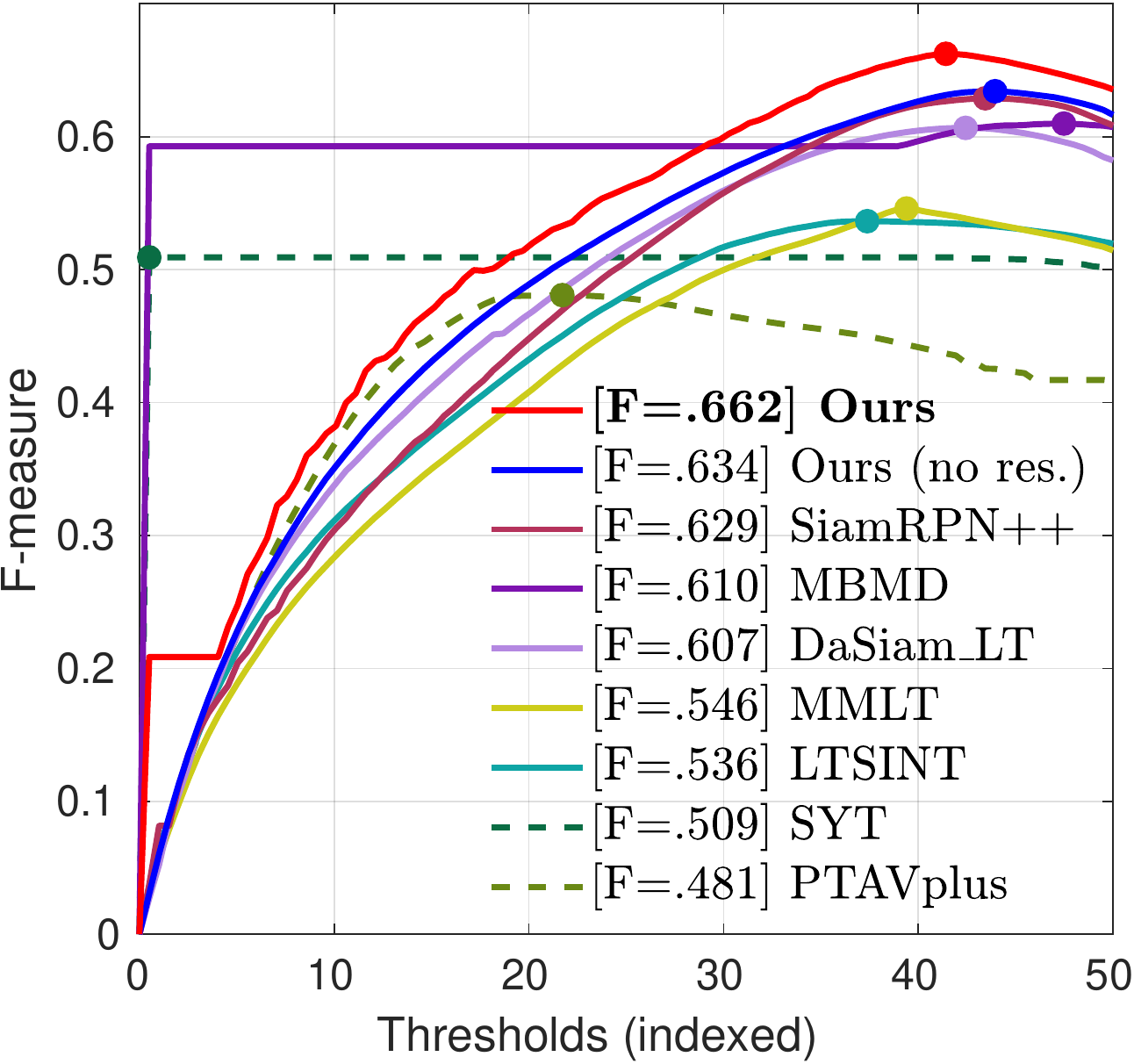}
}
\caption{Results on the VOT 2018 Long-term dataset. The evaluation measures are recall and precision (left) and F-score (right).}
\label{fig:long-term-res}
\end{figure}

\subsection{Long-Term Visual Object Tracking Evaluation}
For evaluating BoLTVOS on long-term visual object tracking, we use the LTB35 dataset \cite{Kristan18ECCVW}. This dataset was used in the VOT 2018 challenge to evaluate long-term tracking performance (VOT18-LT) \cite{Kristan18ECCVW} %
 and consists of 35 videos with a similar setup to many other tracking benchmarks where the first-frame ground truth is given. However, the number of frames per sequence is much larger than for typical tracking benchmarks (4200, compared to 590 for OTB2015 \cite{Wu15TPAMI}). Also in typical tracking benchmarks the object to be tracked is present in almost every frame of the video, whereas for LTB35 on average the object to be tracked disappears and reappears again 12.4 times per video with an average target absence period of 40.6 frames. On this benchmark, trackers need to determine when the target has been lost (or disappeared), and re-detect the target after it is lost. In order to measure this, a tracker must output its detection confidence for each predicted bounding box.

We adopt the standard metrics for LTB35, measuring both the precision and the recall at different detection confidences. %
From the precision (Pr) and recall (Re), the F-score can be calculated as $F = \frac{2PrRe}{(Pr+Re)}$. Trackers are ordered by the maximum $\mathcal{F}$ score they achieve over the different confidence thresholds.
Our results on LTB35 can be found in Figure \ref{fig:long-term-res}. We evaluate both our raw conditional R-CNN bounding box prediction and detection score results (Ours (No Rescoring)), and our results with temporal consistency rescoring (Ours). We compare to the 6 best-performing methods in the 2018 VOT-LT challenge which was won by MBMD (described in \cite{Kristan18ECCVW}). We additionally compare to SiamRPN++ \cite{Li19CVPR}, %
the previously %
 strongest performing method with a maximum $\mathcal{F}$ score of $62.9\%$. 
 
 BoLTVOS outperforms all of the previous methods both with and without rescoring. Without rescoring, we obtain a maximum $\mathcal{F}$ score of $63.4\%$, a score $0.5\%$ higher than the previous best result. With our temporal consistency rescoring algorithm, we achieve %
 a maximum $\mathcal{F}$ score of $66.2\%$. This is $3.3\%$ higher than any previously published method. In this setup, our tracker runs at 1.43 frames per second. %
These strong results on LTB35 show the strength of our method for long-term tracking. Because we are able to detect globally over a whole image, and not just within a local window of the previous detection, we are able to easily re-detect a target after it has disappeared. 

\subsection{Short-Term Visual Object Tracking Evaluation}
For evaluating BoLTVOS on short-term visual object tracking, we use the OTB2015 benchmark \cite{Wu15TPAMI}, one of the most commonly used benchmarks for VOT. This dataset contains 100 videos with an average length of 590 frames. %

\begin{figure}[t]
 \scalebox{1.05}{
 \hspace{-10pt}
\includegraphics[scale=0.226]{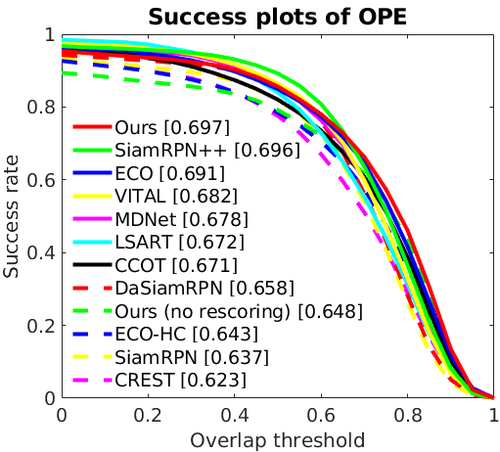}
\includegraphics[scale=0.226]{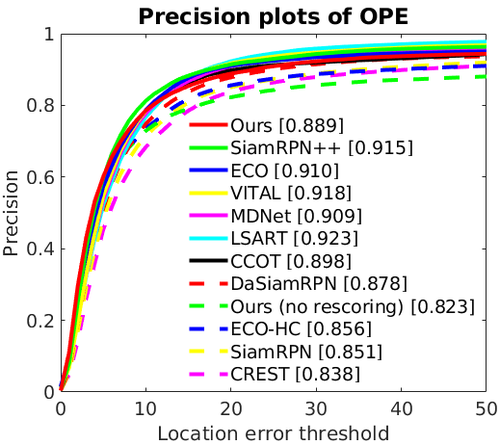}
}
\caption{Success and precision results on the OTB2015 dataset.}
\label{fig:otb-curves}
\end{figure}

We adopt the standard metrics for evaluating on OTB2015 \cite{Wu15TPAMI}, calculating the success and precision of our results over varying overlap thresholds. Methods %
 are ranked by the area under the curve (AUC) of the success curve. %
Fig.~\ref{fig:otb-curves} shows our results both with (Ours), and without (Ours (No Rescoring)) temporal consistency rescoring and compares it to 10 other state-of-the-art tracking approaches:  SiamRPN++ \cite{Li19CVPR}, ECO \cite{Danelljan17CVPR}, VITAL \cite{Song18CVPR}, MDNet \cite{mdnet}, LSART \cite{Sun18CVPR}, C-COT \cite{Danelljan16ECCV}, DaSiamRPN \cite{Zhu18ECCV}, ECO-HC \cite{Danelljan17CVPR}, SiamRPN \cite{Li18CVPRSiamRPN}, and CREST \cite{Song17ICCV}. 

BoLTVOS achieves an AUC score of $69.7\%$, which is slightly better than the previous best result %
by SiamRPN++ \cite{Li19CVPR} of $69.6\%$. Without temporal consistency rescoring, our method achieves %
$64.8\%$, which is competitive with many other state-of-the-art methods even though it uses no temporal consistency cues. In this setup, our tracker runs at 1.43 frames per second.
These strong results show that our BoLTVOS method is not only a powerful method for VOS and long-term VOT, but also shows strong performance for short-term VOT. %

\section{Conclusion}
By dividing the VOS task into box-level tracking and bounding box segmentation, we are able to develop a two-step method, BoLTVOS, that excels in each sub-task separately. 
We present a novel conditional R-CNN and a temporal consistency rescoring algorithm for box-level tracking, which together achieve a new state-of-the-art in long-term tracking on LTB35 and in short term-tracking on OTB2015. By applying Box2Seg on our tracking results, BoLTVOS is able to outperform nearly all other VOS methods on the DAVIS datasets and YouTube-VOS, while being up to 45 times faster than the previous best-performing method and only using the first-frame bounding box.
Our results conclusively show that VOS can benefit from improved box-level tracking. We expect that further progress can be achieved by closer integration of pixel-level segmentation, which currently does not yet feed back into tracking in BoLTVOS.

\footnotesize \PAR{Acknowledgements:} This project has been funded, in parts, by ERC Consolidator Grant DeeViSe (ERC-2017-COG-773161) and by a Google Faculty Research Award. We would like to thank Bo Li for helpful discussions.

{\small
\bibliographystyle{ieee}
\bibliography{abbrev_short,paper}
}

\end{document}

%% file: tables/davis-val.tex
\begin{table}[t]
\centering{}{\footnotesize{}}%
\scalebox{0.92}{
\setlength{\tabcolsep}{3pt}
\begin{tabular}{cccccccc}
\toprule 
 & {\footnotesize{}FT} & {\footnotesize{}M} & {\footnotesize{}$\mathcal{J}$\&$\mathcal{F}$} & {\footnotesize{}$\mathcal{J}$} & {\footnotesize{}$\mathcal{F}$} & {\footnotesize{}$\mathcal{J}_{box}$} & {\footnotesize{}t(s)}\tabularnewline

\midrule
 
{\footnotesize{}Ours} & {\footnotesize{}\ding{55}} & {\footnotesize{}\ding{55}} & {\footnotesize{}$\mathbf{71.9}$} & {\footnotesize{}$\mathbf{68.4}$} & {\footnotesize{}$\mathbf{75.4}$} & {\footnotesize{}$\mathbf{78.5}$} & {\footnotesize{}$1.45$}\tabularnewline

{\footnotesize{}Ours (No Rescoring)} & {\footnotesize{}\ding{55}} & {\footnotesize{}\ding{55}} & {\footnotesize{}$64.9$} & {\footnotesize{}$60.9$} & {\footnotesize{}$68.9$} & {\footnotesize{} 74.5} & {\footnotesize{}$1.45$}\tabularnewline

{\footnotesize{}SiamMask \cite{Wang19CVPR}} & {\footnotesize{}\ding{55}} & {\footnotesize{}\ding{55}} & {\footnotesize{}$55.8$} & {\footnotesize{}$54.3$} & {\footnotesize{}$58.5$} &  {\footnotesize{}$64.3$} & {\footnotesize{}$\mathbf{0.06^{\dagger}}$}\tabularnewline

{\footnotesize{}SiamMask \cite{Wang19CVPR} (Box2Seg)} & {\footnotesize{}\ding{55}} & {\footnotesize{}\ding{55}} & {\footnotesize{}$63.3$} & {\footnotesize{}$59.5$} & {\footnotesize{}$67.3$} &  {\footnotesize{}$64.3$} & {\footnotesize{}$0.11$}\tabularnewline

\midrule

{\footnotesize{}FEELVOS \cite{Voigtlaender19CVPR}} & {\footnotesize{}\ding{55}} & {\footnotesize{}\ding{51}} & {\footnotesize{}$\mathbf{71.5}$} & {\footnotesize{}$\mathbf{69.1}$} & {\footnotesize{}$\mathbf{74.0}$} & {\footnotesize{}$\mathbf{71.4}$} & {\footnotesize{}$0.51$}\tabularnewline

{\footnotesize{}RGMP \cite{Oh18CVPR}} & {\footnotesize{}\ding{55}} & {\footnotesize{}\ding{51}} & {\footnotesize{}$66.7$} & {\footnotesize{}$64.8$} & {\footnotesize{}$68.6$} & {\footnotesize{}66.5} & \textbf{\footnotesize{}$\mathbf{0.28^{\dagger}}$}\tabularnewline

{\footnotesize{}VideoMatch \cite{Hu18ECCV}} & {\footnotesize{}\ding{55}} & {\footnotesize{}\ding{51}} & {\footnotesize{}$62.4$} & {\footnotesize{}$56.5$} & {\footnotesize{}$68.2$} &  {\footnotesize{}$-$} & {\footnotesize{}$0.35$}\tabularnewline

{\footnotesize{}FAVOS \cite{Cheng18CVPR}} & {\footnotesize{}\ding{55}} & {\footnotesize{}\ding{51}} & {\footnotesize{}$58.2$} & {\footnotesize{}$54.6$} & {\footnotesize{}$61.8$} & {\footnotesize{}68.0} & {\footnotesize{}$1.2^{\dagger}$}\tabularnewline

{\footnotesize{}OSMN \cite{Yang18CVPR}} & {\footnotesize{}\ding{55}} & {\footnotesize{}\ding{51}} &  {\footnotesize{}$54.8$} & {\footnotesize{}$52.5$} & {\footnotesize{}$57.1$} & {\footnotesize{}60.1} & \textbf{\footnotesize{}$\mathbf{0.28^{\dagger}}$}\tabularnewline

\midrule

{\footnotesize{}Ours (Fine-tun. Box2Seg)} & {\footnotesize{}\ding{51}} & {\footnotesize{}\ding{51}} & {\footnotesize{}$76.3$} & {\footnotesize{}$72.0$} & {\footnotesize{}$80.6$} & {\footnotesize{}$78.5$} & {\footnotesize{}$\mathbf{1.45}$}\tabularnewline

{\footnotesize{}PReMVOS \cite{Luiten18ACCV}} & {\footnotesize{}\ding{51}} & {\footnotesize{}\ding{51}} & {\footnotesize{}$\mathbf{77.8}$} & {\footnotesize{}$\mathbf{73.9}$} & {\footnotesize{}$\mathbf{81.7}$} & {\footnotesize{}$\mathbf{81.4}$} & {\footnotesize{}$37.6$}\tabularnewline

{\footnotesize{}DyeNet \cite{Li18ECCV}} & {\footnotesize{}\ding{51}} & {\footnotesize{}\ding{51}} & {\footnotesize{}$74.1$} & {\footnotesize{}$-$} & {\footnotesize{}$-$} & {\footnotesize{}$-$} & {\footnotesize{}$9.32^{\dagger}$}\tabularnewline

{\footnotesize{}OSVOS-S \cite{Maninis18TPAMI}} & {\footnotesize{}\ding{51}} & {\footnotesize{}\ding{51}} & {\footnotesize{}$68.0$} & {\footnotesize{}$64.7$} & {\footnotesize{}$71.3$} & {\footnotesize{}$68.4$} &{\footnotesize{}$9^{\dagger}$}\tabularnewline

{\footnotesize{}CINM \cite{Bao18CVPR}} & {\footnotesize{}\ding{51}} &  {\footnotesize{}\ding{51}} & {\footnotesize{}$67.5$} & {\footnotesize{}$64.5$} & {\footnotesize{}$70.5$} & {\footnotesize{}$72.9$} & {\footnotesize{}$>\!120$}\tabularnewline

{\footnotesize{}OnAVOS \cite{voigtlaender17BMVC}} & {\footnotesize{}\ding{51}} & {\footnotesize{}\ding{51}} & {\footnotesize{}$63.6$} & {\footnotesize{}$61.0$} & {\footnotesize{}$66.1$} &  {\footnotesize{}$66.3$} & {\footnotesize{}$26$}\tabularnewline

{\footnotesize{}OSVOS \cite{OSVOS}} & {\footnotesize{}\ding{51}} & {\footnotesize{}\ding{51}} & {\footnotesize{}$60.3$} & {\footnotesize{}$56.6$} & {\footnotesize{}$63.9$} & {\footnotesize{}$57.0$} & {\footnotesize{}$18^{\dagger}$}\tabularnewline

\midrule

{\footnotesize{}GT boxes (Box2Seg)} & {\footnotesize{}\ding{55}} & {\footnotesize{}\ding{51}} & {\footnotesize{}$82.6$} & {\footnotesize{}$79.3$} & {\footnotesize{}$85.8$} & {\footnotesize{}$100.0$} & {\footnotesize{}$-$}\tabularnewline

{\footnotesize{}GT boxes (Fine-tun. Box2Seg)} & {\footnotesize{}\ding{51}} & {\footnotesize{}\ding{51}} & {\footnotesize{}$86.2$} & {\footnotesize{}$81.8$} & {\footnotesize{}$90.5$} &  {\footnotesize{}$100.0$} & {\footnotesize{}$-$}\tabularnewline

\bottomrule
\vspace{-7pt}
\end{tabular}}{\footnotesize{}\caption{\label{tab:results-davis17}Quantitative results on the DAVIS 2017
validation set. FT denotes fine-tuning, M denotes using the first-frame masks, t(s) denotes time per frame
in seconds. $\dagger$: timing extrapolated from DAVIS 2016 assuming
linear scaling in the number of objects.}
}{\footnotesize \par}
\end{table}

%% file: tables/speedplot.tex
\begin{figure}
\centering
\resizebox{\linewidth}{!}{\begin{tikzpicture}[/pgfplots/width=1\linewidth, /pgfplots/height=0.90\linewidth, /pgfplots/legend pos=south east]
    \begin{axis}[ymin=0.50,ymax=0.80,xmin=0.05,xmax=130,enlargelimits=false,
        xlabel=Time per frame (seconds),
        ylabel=Region and contour quality ($\mathcal{J}$\&$\mathcal{F}$),
		font=\small,%
        grid=both,
		grid style=dotted,
        xlabel shift={-2pt},
        ylabel shift={-5pt},
        xmode=log,
        legend columns=1,
        minor ytick={0,0.025,...,1.1},
        ytick={0,0.1,...,1.1},
	    yticklabels={0,.1,.2,.3,.4,.5,.6,.7,.8,.9,1},
	    xticklabels={0.01,.1,1,10,100},
        legend pos= outer north east,
        legend cell align={left}
        ]

	    \addplot[red,mark=*,only marks,line width=0.75, mark size=3.0] coordinates{(1.45,0.719)};
        \addlegendentry{\hphantom{i}Ours}
        
        \addplot[red,mark=+, only marks, line width=0.75, mark size=2.5] coordinates{(0.06,0.558)};
        \addlegendentry{\underline{\hphantom{i}SiamMask \cite{Wang19CVPR}\phantom{iiiiiii}}}        
        
        \addplot[green,mark=+,only marks,line width=0.75, mark size=2.5] coordinates{(0.51,0.715)};
        \addlegendentry{\hphantom{i}FEELVOS \cite{Voigtlaender19CVPR}}
        
        \addplot[green,mark=star, only marks, line width=0.75, mark size=2.5] coordinates{(0.26,0.667)};
        \addlegendentry{\hphantom{i}RGMP \cite{Oh18CVPR}}
        
        \addplot[green,mark=triangle,only marks,line width=0.75, mark size=2.0] coordinates{(0.35,0.624)};
        \addlegendentry{\hphantom{i}VideoMatch \cite{Hu18ECCV}}
        
        \addplot[green,mark=square, only marks, line width=0.75, mark size=2.5] coordinates{(1.2,0.582)};
        \addlegendentry{\hphantom{i}FAVOS \cite{Cheng18CVPR}}  
        
        \addplot[green,mark=Mercedes star, only marks, line width=0.75, mark size=2.5] coordinates{(0.28,0.548)};
        \addlegendentry{\underline{\hphantom{i}OSMN \cite{Yang18CVPR}\phantom{iiiiiiiiiii}}}
        
  	    \addplot[blue,mark=*,only marks,line width=0.75, mark size=3.0] coordinates{(1.45,0.763)};
        \addlegendentry{\hphantom{i}Ours (Fine-tuned)}
        
        \addplot[blue,mark=+,only marks,line width=0.75, mark size=2.5] coordinates{(37.4,0.778)};
        \addlegendentry{\hphantom{i}PReMVOS \cite{Luiten18ACCV, Luiten18DAVIS}}
        
       \addplot[blue,mark=asterisk, only marks, line width=0.75, mark size=2.5]coordinates{(4.66,0.741)};
        \addlegendentry{\hphantom{i}DyeNet \cite{Li18ECCV}}
        
        \addplot[blue,mark=triangle, only marks, line width=0.75, mark size=2.5] coordinates{(9,0.68)};
        \addlegendentry{\hphantom{i}OSVOS-S \cite{Maninis18TPAMI}}
        
        \addplot[blue,mark=square, only marks, line width=0.75, mark size=2.5] coordinates{(120,0.675)};
        \addlegendentry{\hphantom{i}CINM \cite{Bao18CVPR}} 
      
        \addplot[blue,mark=Mercedes star,only marks,line width=0.75, mark size=2.5] coordinates{(26,	0.636)};
        \addlegendentry{\hphantom{i}OnAVOS \cite{voigtlaender17BMVC, voigtlaender17DAVIS}}
        
        \addplot[blue,mark=pentagon, only marks, line width=0.75, mark size=2.5] coordinates{(18,0.603)};
        \addlegendentry{\hphantom{i}OSVOS \cite{OSVOS}}

    \end{axis}
\end{tikzpicture}}
\vspace{-5mm}
   \caption{Quality versus timing on DAVIS 2017. Only SiamMask \cite{Wang19CVPR} and our method (red) are able to work without the ground truth mask of the first frame and require just the bounding box. Methods shown in blue fine-tune on the first-frame mask.}
   \label{fig:qual_vs_time}
\end{figure}

%% file: tables/speedplot_ytbvos.tex
\begin{figure}
\centering
\resizebox{\linewidth}{!}{\begin{tikzpicture}[/pgfplots/width=1\linewidth, /pgfplots/height=0.55\linewidth, /pgfplots/legend pos=south east]
    \begin{axis}[ymin=0.48,ymax=0.73,xmin=0.05,xmax=100,enlargelimits=false,
        xlabel=Time per frame (seconds),
        ylabel=$\mathcal{J}\&\mathcal{F}_{seen+uns.}$,
		font=\small,%
        grid=both,
		grid style=dotted,
        xlabel shift={-2pt},
        ylabel shift={-5pt},
        xmode=log,
        legend columns=1,
        minor ytick={0,0.025,...,1.1},
        ytick={0,0.1,...,1.1},
	    yticklabels={0,.1,.2,.3,.4,.5,.6,.7,.8,.9,1},
	    xticklabels={0.01,.1,1,10,100},
        legend pos= outer north east,
        legend cell align={left}
        ]

	    \addplot[red,mark=*,only marks,line width=0.75, mark size=3.0] coordinates{(1.36,0.657)};
        \addlegendentry{\hphantom{i}Ours}
        
        \addplot[red,mark=+, only marks, line width=0.75, mark size=2.5] coordinates{(0.06,0.528)};
        \addlegendentry{\underline{\hphantom{i}SiamMask \cite{Wang19CVPR}\phantom{iiiiiii}}}        
        
        \addplot[green,mark=star, only marks, line width=0.75, mark size=2.5] coordinates{(0.26,0.538)};
        \addlegendentry{\underline{\hphantom{i}RGMP \cite{Oh18CVPR}\phantom{iiiiiiiiiii}}}
        
   	    \addplot[blue,mark=*,only marks,line width=0.75, mark size=3.0] coordinates{(1.36,0.711)};
        \addlegendentry{\hphantom{i}Ours (Fine-tuned)}
        
        \addplot[blue,mark=+,only marks,line width=0.75, mark size=2.5] coordinates{(6,0.669)};
        \addlegendentry{\hphantom{i}PReMVOS \cite{Luiten18ACCV, Luiten18ECCVW}}
        
        \addplot[blue,mark=Mercedes star,only marks,line width=0.75, mark size=2.5] coordinates{(24.5,	0.552)};
        \addlegendentry{\hphantom{i}OnAVOS \cite{voigtlaender17BMVC, voigtlaender17DAVIS}}
        
        \addplot[blue,mark=pentagon, only marks, line width=0.75, mark size=2.5] coordinates{(17,0.588)};
        \addlegendentry{\hphantom{i}OSVOS \cite{OSVOS}}

    \end{axis}
\end{tikzpicture}}
\vspace{-5mm}
   \caption{Quality versus timing on YouTube-VOS \cite{Xu18ECCV}. %
   }
   \label{fig:speedplot-ytbvos}
   \vspace{-1mm}
\end{figure}

%% file: tables/davis16.tex
\begin{table}[t]
\centering{}{\footnotesize{}}%
\scalebox{0.92}{
\setlength{\tabcolsep}{3pt}
\begin{tabular}{cccccccc}
\toprule 
 & {\footnotesize{}FT} & {\footnotesize{}M} & {\footnotesize{}$\mathcal{J}$\&$\mathcal{F}$} & {\footnotesize{}$\mathcal{J}$} & {\footnotesize{}$\mathcal{F}$} & {\footnotesize{}$\mathcal{J}_{box}$} & {\footnotesize{}t(s)}\tabularnewline
\midrule

{\footnotesize{}Ours} & {\footnotesize{}\ding{55}} & {\footnotesize{}\ding{55}} & \textbf{\footnotesize{}$\mathbf{79.6}$} & {\footnotesize{}$\mathbf{78.1}$} & {\footnotesize{}$\mathbf{81.2}$} & {\footnotesize{}$\mathbf{88.3}$} & {\footnotesize{}$0.72$}\tabularnewline

{\footnotesize{}Ours (No Rescoring)} & {\footnotesize{}\ding{55}} & {\footnotesize{}\ding{55}} &  \textbf{\footnotesize{}$\mathbf{79.6}$} & {\footnotesize{}$\mathbf{78.1}$} & {\footnotesize{}$81.1$} & {\footnotesize{}$\mathbf{88.3}$} & {\footnotesize{}$0.72$}\tabularnewline

{\footnotesize{}SiamMask \cite{Wang19CVPR}} & {\footnotesize{}\ding{55}} & {\footnotesize{}\ding{55}} & \textbf{\footnotesize{}$69.8$} & {\footnotesize{}$71.7$} & {\footnotesize{}$67.8$} & {\footnotesize{}$73.3$} & {\footnotesize{}$\mathbf{0.03}$}\tabularnewline

{\footnotesize{}SiamMask \cite{Wang19CVPR} (Box2Seg)} & {\footnotesize{}\ding{55}} & {\footnotesize{}\ding{55}} & {\footnotesize{}$75.9$} & {\footnotesize{}$75.6$} & {\footnotesize{}$76.3$} & {\footnotesize{}$73.3$} & {\footnotesize{}$0.06$}\tabularnewline

\midrule

{\footnotesize{}RGMP \cite{Oh18CVPR}} & {\footnotesize{}\ding{55}} & {\footnotesize{}\ding{51}} & \textbf{\footnotesize{}$\mathbf{81.8}$} & \textbf{\footnotesize{}$81.5$} & {\footnotesize{}$82.0$} & {\footnotesize{}$79.3$} & \textbf{\footnotesize{}$\mathbf{0.14}$}\tabularnewline

{\footnotesize{}FEELVOS \cite{Voigtlaender19CVPR}} & {\footnotesize{}\ding{55}} & {\footnotesize{}\ding{51}} & \textbf{\footnotesize{}$81.7$} &{\footnotesize{}$81.1$} & {\footnotesize{}$\mathbf{82.2}$} & {\footnotesize{}$80.2$} & {\footnotesize{}$0.45$}\tabularnewline

{\footnotesize{}FAVOS \cite{Cheng18CVPR}} & {\footnotesize{}\ding{55}} & {\footnotesize{}\ding{51}} &  {\footnotesize{}$81.0$} & {\footnotesize{}$\mathbf{82.4}$} & {\footnotesize{}$79.5$} & {\footnotesize{}$\mathbf{83.1}$} & {\footnotesize{}$0.6$}\tabularnewline

{\footnotesize{}VideoMatch \cite{Hu18ECCV}} & {\footnotesize{}\ding{55}} & {\footnotesize{}\ding{51}} & {\footnotesize{}$80.9$} & {\footnotesize{}$81.0$} & {\footnotesize{}$80.8$} & {\footnotesize{}$-$} & {\footnotesize{}$0.32$}\tabularnewline

{\footnotesize{}PML \cite{Chen18CVPR}} & {\footnotesize{}\ding{55}} & {\footnotesize{}\ding{51}} & {\footnotesize{}$77.4$} & {\footnotesize{}$75.5$} & {\footnotesize{}$79.3$} & {\footnotesize{}$75.9$} & {\footnotesize{}$0.28$}\tabularnewline

{\footnotesize{}OSMN \cite{Yang18CVPR}} & {\footnotesize{}\ding{55}} & {\footnotesize{}\ding{51}} & {\footnotesize{}$73.5$} & {\footnotesize{}$74.0$} & {\footnotesize{}$72.9$} & {\footnotesize{}$71.8$} & {\footnotesize{}$\mathbf{0.14}$}\tabularnewline

\midrule 

{\footnotesize{}Ours (Fine-tun. Box2Seg)} & {\footnotesize{}\ding{51}} & {\footnotesize{}\ding{51}} &  {\footnotesize{}$\mathbf{87.7}$} & {\footnotesize{}$\mathbf{86.3}$} & {\footnotesize{}$\mathbf{89.1}$} & {\footnotesize{}$88.3$} & {\footnotesize{}$\mathbf{0.72}$}\tabularnewline

{\footnotesize{}PReMVOS \cite{Luiten18ACCV}} & {\footnotesize{}\ding{51}} & {\footnotesize{}\ding{51}} & {\footnotesize{}$86.8$} & {\footnotesize{}$84.9$} & {\footnotesize{}$88.6$} & {\footnotesize{}$\mathbf{89.9}$} & {\footnotesize{}$32.8$}\tabularnewline

{\footnotesize{}DyeNet \cite{Li18ECCV}} & {\footnotesize{}\ding{51}} & {\footnotesize{}\ding{51}} & {\footnotesize{}$-$} & {\footnotesize{}$86.2$} & {\footnotesize{}$-$} & {\footnotesize{}$-$} & {\footnotesize{}$4.66$}\tabularnewline

{\footnotesize{}OSVOS-S \cite{Maninis18TPAMI}} & {\footnotesize{}\ding{51}} & {\footnotesize{}\ding{51}} & {\footnotesize{}$86.5$} & {\footnotesize{}$85.6$} & {\footnotesize{}$87.5$} & {\footnotesize{}$84.4$} & {\footnotesize{}$4.5$}\tabularnewline

{\footnotesize{}OnAVOS \cite{voigtlaender17BMVC}} & {\footnotesize{}\ding{51}} & {\footnotesize{}\ding{51}} & {\footnotesize{}$85.0$} & {\footnotesize{}$85.7$} & {\footnotesize{}$84.2$} & {\footnotesize{}$84.1$} & {\footnotesize{}$13$}\tabularnewline

{\footnotesize{}CINM \cite{Bao18CVPR}} & {\footnotesize{}\ding{51}} & {\footnotesize{}\ding{51}} & {\footnotesize{}$84.2$} & {\footnotesize{}$83.4$} & {\footnotesize{}$85.0$} & {\footnotesize{}$83.6$} & {\footnotesize{}$>120$}\tabularnewline

{\footnotesize{}OSVOS \cite{OSVOS}} & {\footnotesize{}\ding{51}} & {\footnotesize{}\ding{51}} & {\footnotesize{}$80.2$} & {\footnotesize{}$79.8$} & {\footnotesize{}$80.6$} & {\footnotesize{}$76.0$} & {\footnotesize{}$9$}\tabularnewline

\midrule

{\footnotesize{}GT boxes (Box2Seg)} & {\footnotesize{}\ding{55}} & {\footnotesize{}\ding{51}} & \textbf{\footnotesize{}$80.5$} & {\footnotesize{}$79.1$} & {\footnotesize{}$81.9$} & {\footnotesize{}$100.0$} & {\footnotesize{}$-$}\tabularnewline

{\footnotesize{}GT boxes (Fine-tun. Box2Seg)} & {\footnotesize{}\ding{51}} & {\footnotesize{}\ding{51}} & {\footnotesize{}$89.0$} & {\footnotesize{}$87.6$} & {\footnotesize{}$90.5$} & {\footnotesize{}$100.0$} & {\footnotesize{}$-$}\tabularnewline

\bottomrule
\end{tabular}}{\footnotesize{}\caption{\label{tab:results-davis16}Quantitative results on the DAVIS 2016
validation set. FT denotes fine-tuning, M denotes using the first-frame mask, and t(s) denotes time per frame
in seconds.}
}
\end{table}

%% file: tables/youtubevos.tex
\begin{table}[t]
\centering{}{\footnotesize{}}%
\scalebox{0.92}{
\setlength{\tabcolsep}{3pt}
\begin{tabular}{cccccccc}
\toprule 
 & {\footnotesize{}FT} & {\footnotesize{}M} & {\footnotesize{}$\mathcal{J}\&\mathcal{F}_{seen+uns.}$} & {\footnotesize{}$\mathcal{J}_{seen}$} & {\footnotesize{}$\mathcal{J}_{unseen}$} & {\footnotesize{}t(s)}\tabularnewline
\midrule

{\footnotesize{}Ours} & {\footnotesize{}\ding{55}} & {\footnotesize{}\ding{55}} & {\footnotesize{}$\mathbf{65.7}$} & {\footnotesize{}$\mathbf{67.3}$} & \textbf{\footnotesize{}$\mathbf{58.9}$} & {\footnotesize{}$1.36$}\tabularnewline

{\footnotesize{}Ours (No Rescoring)} & {\footnotesize{}\ding{55}} & {\footnotesize{}\ding{55}} & {\footnotesize{}$61.2$} & {\footnotesize{}$61.9$} & \textbf{\footnotesize{}$54.8$} & {\footnotesize{}$1.36$}\tabularnewline

{\footnotesize{}SiamMask \cite{Wang19CVPR}} & {\footnotesize{}\ding{55}} & {\footnotesize{}\ding{55}} & {\footnotesize{}$52.8$} & {\footnotesize{}$60.2$} & \textbf{\footnotesize{}$45.1$} &  {\footnotesize{}$\mathbf{0.06}$}\tabularnewline

\midrule

{\footnotesize{}RGMP \cite{Oh18CVPR}} & {\footnotesize{}\ding{55}} & {\footnotesize{}\ding{51}} & \textbf{\footnotesize{}$53.8$} & {\footnotesize{}$59.5$} & \textbf{\footnotesize{}$45.2$} & \textbf{\footnotesize{}$0.26$}\tabularnewline

\midrule 

{\footnotesize{}Ours (Fi.-tu. Box2Seg)} & {\footnotesize{}\ding{51}} & {\footnotesize{}\ding{51}} & {\footnotesize{}$\mathbf{71.1}$} & {\footnotesize{}$\mathbf{71.6}$} & {\footnotesize{}$\mathbf{64.3}$} & {\footnotesize{}$\mathbf{1.36}$}\tabularnewline

{\footnotesize{}PReMVOS \cite{Luiten18ACCV,Luiten18ECCVW}} & {\footnotesize{}\ding{51}} & {\footnotesize{}\ding{51}} & {\footnotesize{}$66.9$} & {\footnotesize{}$71.4$} & {\footnotesize{}$56.5$} & {\footnotesize{}$6$}\tabularnewline

{\footnotesize{}OnAVOS \cite{voigtlaender17BMVC}} & {\footnotesize{}\ding{51}} & {\footnotesize{}\ding{51}} & {\footnotesize{}$55.2$} & {\footnotesize{}$60.1$} & {\footnotesize{}$46.6$} & {\footnotesize{}$24.5$}\tabularnewline

{\footnotesize{}OSVOS \cite{OSVOS}} & {\footnotesize{}\ding{51}} & {\footnotesize{}\ding{51}} & {\footnotesize{}$58.8$} & {\footnotesize{}$59.8$} & {\footnotesize{}$54.2$} & {\footnotesize{}$17$}\tabularnewline

\bottomrule
\end{tabular}}{\footnotesize{}\caption{\label{tab:results-youtubevos}Quantitative results on the YouTube-VOS \cite{Xu18ECCV} validation set. FT denotes fine-tuning, M denotes using the first-frame mask, and t(s) denotes time per frame
in seconds.}
}
\end{table}